\documentclass[conference]{IEEEtran}
\ifCLASSINFOpdf
  \usepackage[pdftex]{graphicx}
\else
  \usepackage[dvips]{graphicx}
\fi
\begin{document}
%
\title{Fast, Smart Neuromorphic Sensors Based on Heterogeneous Networks 
and Mixed Encodings}

\author{\IEEEauthorblockN{Angel Yanguas-Gil}
\IEEEauthorblockA{Energy Systems Division\\
Argonne National Laboratory\\
Lemont, Illinois 60439\\
Email: ayg@anl.gov}}


%


\maketitle

\begin{abstract}
Neuromorphic architectures are ideally suited for the implementation of smart sensors able
to react, learn, and respond to a changing environment. Our work uses the insect brain
as a model to understand how heterogeneous architectures, incorporating different types
of neurons and encodings, can be leveraged to create systems integrating input processing,
evaluation, and response. Here we show how the combination of time and rate encodings
can lead to fast sensors that are able
to generate a hypothesis on the input in only a few cycles and then use 
that hypothesis as secondary input for more detailed analysis.
\end{abstract}


%
\IEEEpeerreviewmaketitle

\section{Introduction}

Smart sensors is one of the areas in which neuromorphic computing can be transformational: the ability
to build compact systems able to react, learn, adapt, communicate, and remember while immersed in a changing, unpredictable environment could find applications across many domains.
Insects provide an ideal model system for the exploration of neuromorphic
architectures that can act as smart sensors.  A species such as Drosophila
is able to process and integrate multiple inputs,
exhibit short and long term memory, and display a variety of learning behaviors, all
while being proficient while airborne. 
This range of behavior is achieved with a comparatively small number of neurons,
100,000 in the case of Drosophila,\cite{Chiang_drosophilabrain_2011} and fast response times. For instance,
 olfactory receptor neurons
can resolve input fluctuations at more than 100 Hz.\cite{Szyszka_speedolfactive_2014}

It is not unreasonable to assume that, should we had a 
better understanding of the design principles underlying the insect's brain, the fabrication
of chips implementing such complex functionality would be within the reach of our current device fabrication capabilities.
The possibility of implementing complex systems with a relatively few neurons also broadens the range
of platforms that could be employed to manufacture such sensors. Examples include large ($\ge 90$ nm) technology
 nodes or wide bandgap semiconductors such as SiC for operation
under extreme environments, and the use of TFT or
printing and additive manufacturing technologies.\cite{Fortunato_TFTelectronics_2012, Neudeck_SiCVenus_2016}

\begin{figure}[!t]
\centering
\includegraphics[width=2.5in]{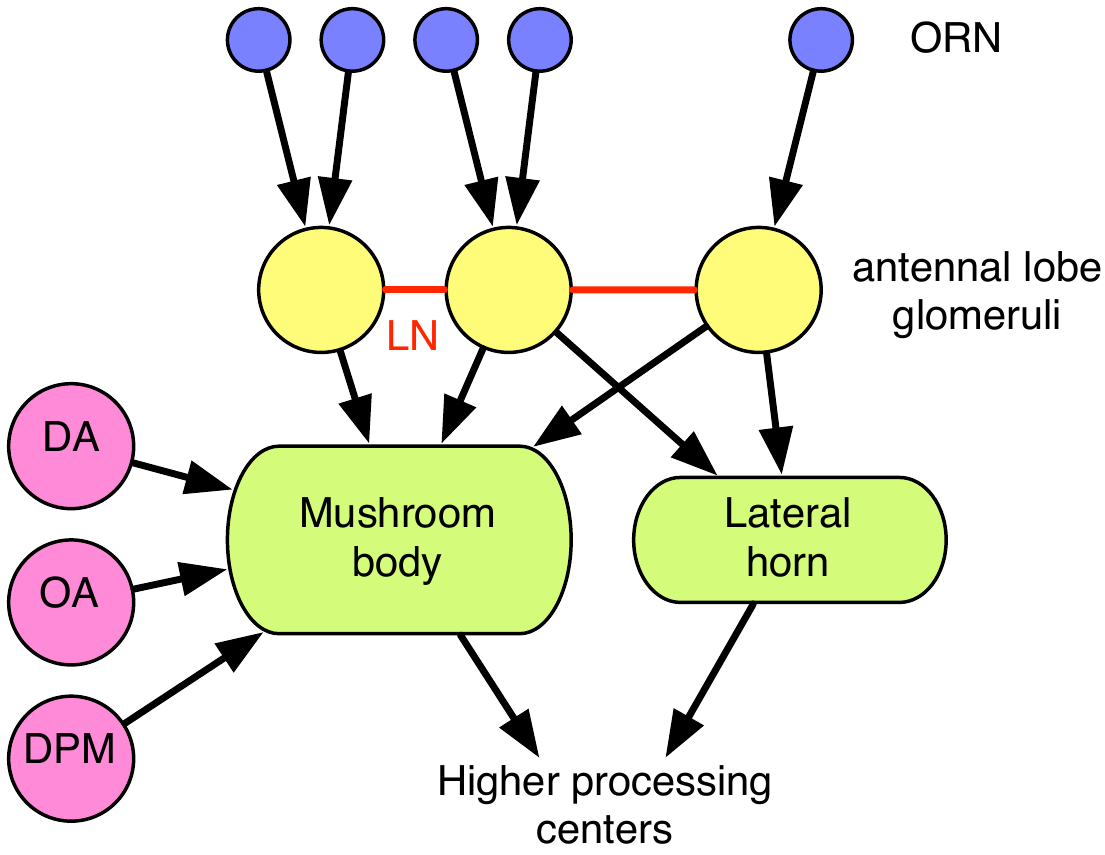}
\caption{Scheme of the olfactive processing pathway on higher insects. Despite its small size, the arthropod brain
is highly modular: inputs from the antennal lobes project
into two separate pathways: the lateral horn, responsible of innate responses, and the mushroom body, one of
the key centers for memory and learning in the insects. Learning in the mushroom body is modulated by 
independent modulatory circuits. }
\label{fig_fly}
\end{figure}

However, despite having
various orders of magnitude fewer neurons than mammals, the insect brain is remarkably heterogeneous. It
 is is composed
of a number of well-differentiated
functional structures, with clear parallelisms with the much larger mammal brain. \cite{Strausfeld_arthopodvertebrate_2013}
An example is provided
in Figure \ref{fig_fly}, where we show a functional model of the olfactory processing and learning subsystem of
\emph{Drosophila melanogaster}.
It also exhibits multiple types of neurons, neurotransmitters and neuromodulators,
 \cite{Chiang_drosophilabrain_2011} as well as multiple encodings: population, sparse, spike timing,
 and single spike codes have all been experimentally observed. \cite{Joesch_motionsensitivefly_2008,PerezOrive_odorrepresentation_2002}
Recent studies also suggest that dynamic plasticity can be associated with a small subset of neurons, providing yet
another source of heterogeneity.\cite{aso_mushroom_2014, aso_neuronal_2014}

Understanding the behavior of such heterogeneous systems, and in particular
the way in which different components can be integrated to achieve a desired functionality, can
help accelerate the development of standalone neuro-inspired smart sensors, as well as neuromorphic systems in general. 
This knowledge can also inform the design and fabrication of neuromorphic systems from
a hardware perspective: the choices that we make building such a system, from standard vs custom ASIC
chips down to spiking vs non-spiking, analog vs digital, CMOS vs memristive-based, and standard vs novel materials
choices in novel architectures, can have
a strong impact in the ultimate performance and capabilities in ways that are not yet fully understood.


This manuscript is structured as follows: in Section \ref{sec_model}, we introduce our approach to integrate spiking neurons
with non-spiking recurrent neurons, and we introduce the types of architectures that are the focus of our study. In Section \ref{sec_validation} we validate that the state based representation of spiking neurons provides an accurate description of
spiking neurons. We then use this model in Section \ref{sec_encodings} to explore different types of encodings,
and compare their pattern recognition capabilities using the same network. In Sections \ref{sec_mushroom} and
\ref{sec_modulation} we focus on the sparse coding approach enabled by the mushroom body type of architectures. We also show
how this type of architecture provides the ideal substrate for implementing online modulatory learning.
Finally, we discuss the implications that our results have on the design and choice of neuromorphic hardware.

\section{Model}
\label{sec_model}

\subsection{Integration of spiking and non-spiking neural networks}

A central aspect of our model is the integration of spiking and non-spiking neural neurons. Our approach
was to develop a synchronous, recurrent, yet accurate representation of a leaky integrate and fire model. 

Our starting point is the standard leaky integrate and fire model:\cite{Brunel_LIFdynamics_2000}
\begin{equation}
\tau \frac{d v_i}{dt} = - v_i + \sum_j w_{ji} \delta(t-t_j) + v_\mathrm{ext}^{(i)}
\label{eq_spike}
\end{equation}
subject to the spiking condition:
\begin{equation}
v_i(t) = 1 \Rightarrow v_i(t + t') = 0 \hspace{0.5cm} \forall \hspace{0.5cm} t' < \tau_r
\end{equation}
where $\tau_r$ is an absolute refractory period.

We can take advantage of the fact that for any $\Delta t \le \tau_r$ we can have at least one spike to
integrate over $\Delta t$ and express the output of each neuron as a binary variable $s_i = 0,1$
at each interval. In order to transform the firing times $t_j$ of the neurons projecting into 
our neuron into time steps $n_i$, so that $n_j\Delta t  \ge t_j < (n_j+1)\Delta t$, we assume that $t_j$
is homogeneously distributed within that interval:  $P(t_j) = \frac{1}{\Delta t}$. This approximation,
which is equivalent to assuming that we lose any information below our time scale $\Delta t$,
allows us to formulate Eq. \ref{eq_spike} as:
\begin{equation}
\label{eq_state}
\begin{array}{lcl}
v_i(n) & = &(1-s_i(n-1)) \left[\xi_i(n)\left(1-e^{-1/\tau_m}\right)\right.  \\
& & + \left. v_i(n-1)e^{-1/\tau_m} \right]  \\
&  & + s_i(n-1)\xi_i(n)\left(1- \tau_m (1- e^{-1/\tau_m})\right)  \\
s_i(n) & = &  H(v_i(n)-v_{0i}) \\
\xi_i(n) & = & v_\mathrm{ext}^{(i)} + \sum_j \frac{w_{ji} }{\tau_r} s_j(n-n_{ji})
\end{array}
\end{equation}

Here $s_i = 0,1$ depending on whether the neuron is spiking on a given time step,
$v_i$ is the membrane potential, $\tau_m$ is the neuron time constant normalized to the
absolute refractory period, and $H(\cdot)$ is the Heaviside or step function. The presence of two terms,
one proportional to $(1-s_i)$ and another to $s_i$, in Eq. \ref{eq_state} comes from the need of considering the
impact of the absolute refractory period on the evolution of the membrane potential $v_i$:
if the neuron spiked in the prior interval, the increase in potential will be
lower for the same total input since the neuron is active only during a fraction of the time.

If we neglect the impact of the refractory period
that we just mentioned from Eq. \ref{eq_state}, the recurrence relation for $v_i$ simplifies to:
\begin{equation}
\label{eq_statesimp}
\begin{array}{lcl}
v_i(n) & = & \left(1-e^{-1/\tau_m}\right) \xi_i(n) \\
& & + \left( (1-s_i(n-1)\right) v_i(n-1)e^{-1/\tau_m} \\
s_i(n) & = &  H(v_i(n)-v_{0i}) \\
\xi_i(n) & = & v_\mathrm{ext}^{(i)} + \sum_j \frac{w_{ji} }{\tau_r} s_j(n-n_{ji})
\end{array}
\end{equation}
Eq. \ref{eq_statesimp} represents a recurrent version of the classic McCulloch-Pitts model.
\cite{McCulloch_neuronmodel_1943}  If we
can neglect the memory effect on the neuron the classic model is recovered.

We can use Eq. \ref{eq_state} as a starting point to implement a training algorithm based
on the stochastic gradient descent method.  In order to do that, 
we simply replace the Heaviside function in Eq.
\ref{eq_state}  by
a differentiable activation function $a_i$. Here, we use the logistic function $\sigma(\cdot)$
weighed by a factor $\beta$ used to control the sharpness of the transition, so that:
\begin{equation}
a_i = \sigma(\beta( v_i - v^{(i)}_0))
\end{equation}
The resulting equation is non-linear in $a_i$ due to the presence of the $(1-s_i)$ and $s_i$ factors mentioned in the previous
section.

\subsection{Architecture}

\begin{figure}[!t]
\centering
\includegraphics[width=2.5in]{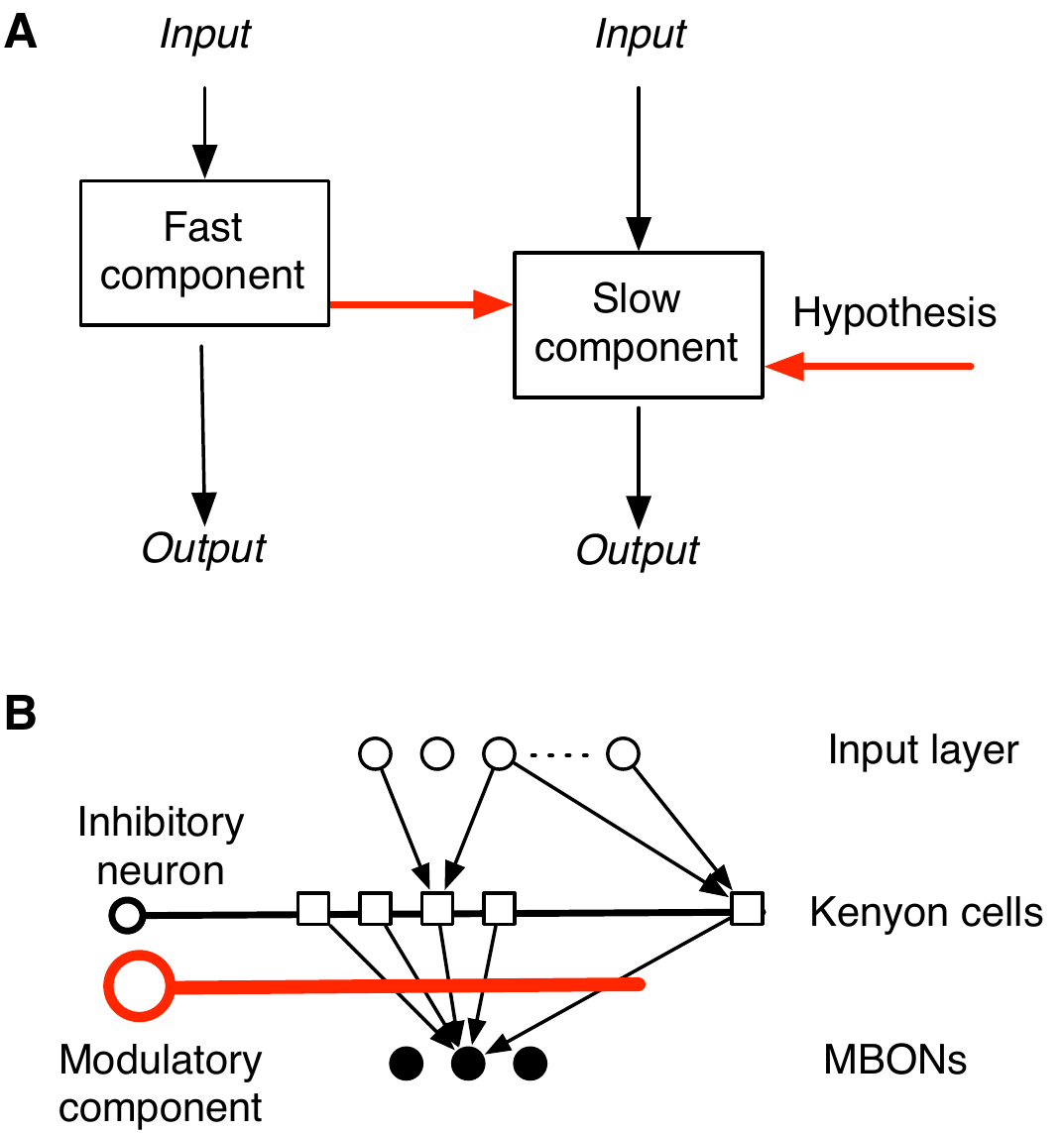}
\caption{Scheme of the olfactive processing pathway on a higher insects. Inputs from the antennal lobes project
into two separate pathways: the lateral horn, responsible of innate responses, and the mushroom body, one of
the key centers for memory and learning in the insects. The mushroom body is modulated by modulatory circuits. }
\label{fig_architecture}
\end{figure}

In this work we have focused on the architecture shown in Figure \ref{fig_architecture}.
This design is clearly inspired on one of
the  memory and learning centers in the insect brain shown in Figure \ref{fig_fly}.
\cite{aso_mushroom_2014, hige_heterosynaptic_2015}

We consider systems composed of two parallel pathways for input processing [Figure \ref{fig_architecture}(A)]. A \emph{fast component}
contains innate information that is processed using a neural network based on either single spike, spike timing, or rate encoding. This
structure is an abstraction from the lateral horn in the insect brain. 


In a second pathway, we have a fan-out fan-in network for sparse encoding of information [Figure \ref{fig_architecture}(B)].
\cite{Garcia-Sanchez2003} The input
data projects into a set of identical neurons (Kenyon cells) that are sparsely connected.
The connectivity is randomly determined so that
only a fraction of input neurons feed into each Kenyon cells. These cells can be recurrently connected. The sparsity of neural activity
is maintained through a single inhibitory neuron that receives input from and connects to every Kenyon cell. The Kenyon cells project
into an output layer (mushroom body output neurons - MBONs). Only the synaptic weights between the Kenyon cells and
the output neurons are plastic.\cite{hige_heterosynaptic_2015}
 The rest of synapses are kept constant during the offline or online training.

\subsection{Modulation as enabler of state or hypothesis-dependent learning}
\label{sec_modulation}

A key component of biological systems is the ability to decide when to learn and when not to learn. In the insect brain, clusters
of modulatory neurons implement such capability though modulatory learning.\cite{aso_mushroom_2014, aso_neuronal_2014}
If in spike timing dependent plasticity the synaptic strength $W_\mathrm{AB}$
between two neurons A and B is a function of their activity: $W_\mathrm{AB} = f(\mathrm{A}, \mathrm{B})$, in the case
of modulated learning, the synaptic strength is a function of the pre-synaptic neuron and a \emph{third} neuron C:
$W_\mathrm{AB} = f(\mathrm{A}, \mathrm{C})$, and optionally the post-synaptic neuron as well. This provides a 
substrate to integrate logic and memory through dynamic plasticity: we can actuate learning or strengthen certain
pathways by the action of a third input. Details on how to implement
such functionality from a hardware perspective will be shown elsewhere.

\section{Results}

\subsection{The state model of spiking neurons is functionally equivalent to the LIF model for sparse activity}
\label{sec_validation}

\begin{figure}[!t]
\centering
\includegraphics[width=3in]{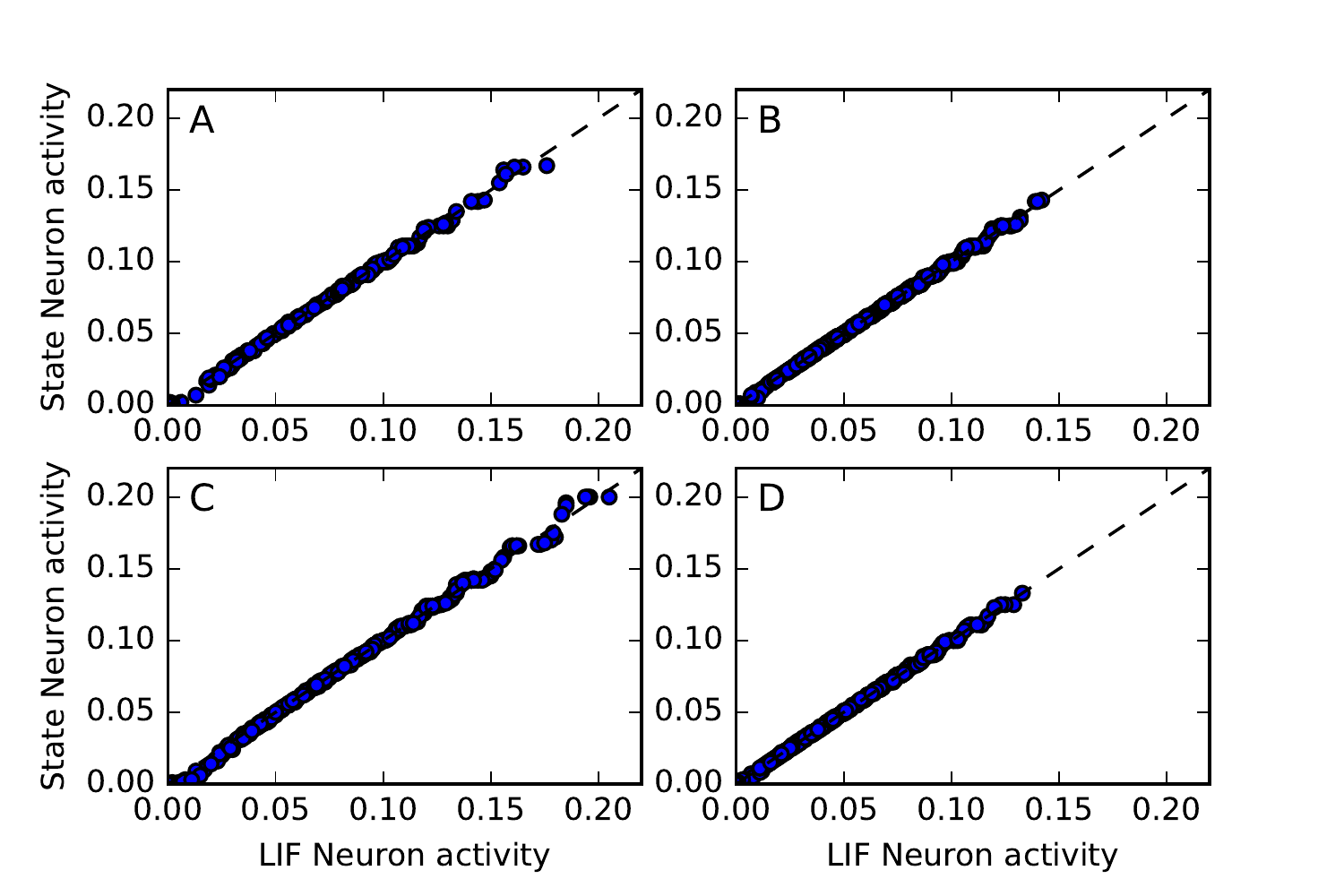}
\caption{Correlation between the standard LIF model and the state model of a spiking neuron used
in this work}
\label{fig_activity}
\end{figure}

In order to understand the impact of our synchronous approximation to a leaky integrate and fire model, we have compared
the activity pattern of our model and asynchronous model for a series of randomly connected networks subject to a random
set of inputs. In Figure \ref{fig_activity} we show the correlation between the activity of a leaky integrate and fire modeled following
equation \ref{eq_spike} and its corresponding state representation given by Eq. \ref{eq_state}. .

The correlation between the two models is excellent, indicating that the loss of information of the exact firing times
does not negatively impact its functionality for a broad set of conditions. In particular, this allows us to extract the 
lessons from prior exhaustive studies on the leaky integrate and fire neurons to a recurrent, synchronous model.\cite{Brunel_LIFdynamics_2000} 

\subsection{Comparison between different encodings: MNIST benchmarks}
\label{sec_encodings}

\begin{figure}[!t]
\centering
\includegraphics[width=2.5in]{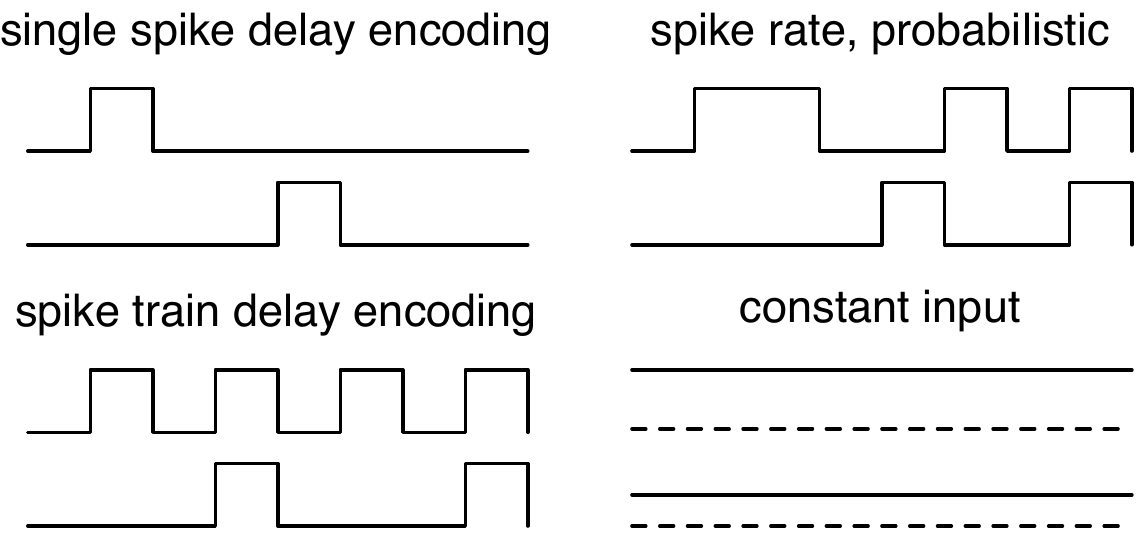}
\caption{Different encodings implemented in this work}
\label{fig_encoding}
\end{figure}

In order to explore the ability of different encodings to learn and discriminate between different inputs we have used the 
state model introduced above to benchmark the performance of a shallow network against the MNIST dataset. 
We have considered the following
four types of encoding (Figure \ref{fig_encoding}):
\begin{enumerate}
\item \emph{Single spike delay encoding}. In this case the input is codified as a single spike whose delay with respect to a
common epoch decreases with the input intensity.

\item \emph{Spike train delay encoding}. The single spikes from the time encoding are subsequently followed by a periodic train
of spikes spaced by an interval equal to the time delay.

\item \emph{spike rate, probabilistic}. The spiking neurons are fed a constant input proportional to the intensity of each pixel in the input
data.

\item \emph{Constant analog input}. The input is codified as constant input signal.

\end{enumerate}

For each of the proposed encodings, we applied a stochastic gradient descent method to optimize our networks
using a cost function that optimized the total number of spikes:
\begin{equation}
C = \frac{1}{N} \sum_i\left(\sum_m a_i(m)-n_i\right)^2
\end{equation}
regardless of when those spikes took place. 

\begin{figure}[!tb]
\centering
\includegraphics[width=3in]{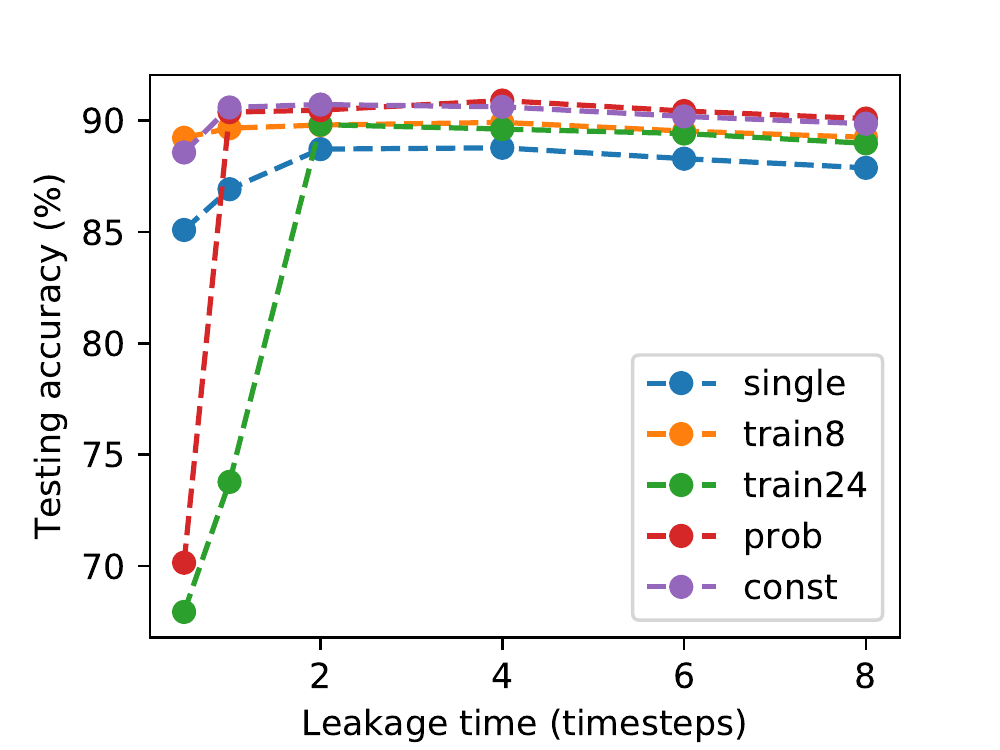}
\caption{MNIST accuracy of a shallow spiking neural network as a function of the neuron leakage time for various encodings.
single: single spike delay; train8: spike train delay encoding, 8 time steps; train24: spike train delay encoding, 24 time steps;
prob: probabilistic spike rate; const: constant input }
\label{fig_spiking}
\end{figure}

In Figure \ref{fig_spiking} we show the testing accuracy for the encodings considered after being trained against 180,000
randomly selected images from MNIST's training set. The results are shown as a function of
the characteristic time of the neurons $\tau_m$. Low values of $\tau_m$ correspond to short-term memory neurons that emphasizes
the contributions within a given interval.  The results show testing accuracies as high as 90\%. This compares well with state of
the art results for shallow (no hidden layers) neural networks.\cite{Lecun_MNIST_1998}
In particular, if we optimize a shallow network of non-spiking
sigmoidal neurons using the same cost function and stochastic gradient descent method, the resulting accuracy is 91\%.

\begin{table}[!t]
\caption{MNIST test accuracy: comparison between sigmoidal and binary outputs for multiple encodings}
\label{table_encoding}
\centering
\begin{tabular}{|c||c|c|c|}
\hline
Encoding & Sigmoidal & Binary non-excl. & Binary excl.\\
\hline
\hline
Single spike delay & 88\% & 83\% & 65\% \\
\hline
Spike train delay & 90\% & 90\% & 87\% \\
\hline
Spike rate, prob & 91\% & 91\% & 87\% \\
\hline
Constant & 88\% & 83\% & 65\% \\
\hline
\end{tabular}
\end{table}

Overall, the conclusions of this analysis is that the learning capability of spiking neurons is similar to that of non-spiking
neurons, at least in the context of the MNIST dataset. Moreover, even in the case of a spike rate, the testing accuracy is
fairly insensitive to the sampling length. In this particular case, a testing accuracy of 90\% is achieved in only
eight steps. 

The results obtained in Figure \ref{fig_spiking} are obtained by changing the binary output of the neuron by a
steep sigmoidal function.
If we switch back to a binary output, the differences between the different encodings become more pronounced. The results are summarized
in Table \ref{table_encoding}. The non-exclusive accuracy column
includes all cases where the correct digit achieves the maximum number of spikes, regardless of whether other digits also achieves the
same number of spikes. In contrast, the exclusive accuracy column considers only those cases where no other digits achieve the
maximum number of spikes.  The need to distinguish between these to cases comes from the fact that the output is measured in terms
of the number of spikes, so two neurons can have identical outputs. According to the results obtained, the single spike encoding's testing (exclusive) accuracy
drops below 70\%. In contrast, both the spike train time delay and the probabilistic spike rate encodings maintain
a high testing accuracy with binary outputs.

\subsection{Sparse encoding using mushroom body architectures}
\label{sec_mushroom}

 
The fan-out fan-in architecture of the mushroom body shown in Figure \ref{fig_architecture} casts the input into a
higher dimensional space, with learning taking
place solely in the fan-in part of the network. The number of Kenyon cells $N$ and the number random connections $n_c$ of
the Kenyon cells are the two key structural variables.

\begin{figure}[!t]
\centering
\includegraphics[width=3in]{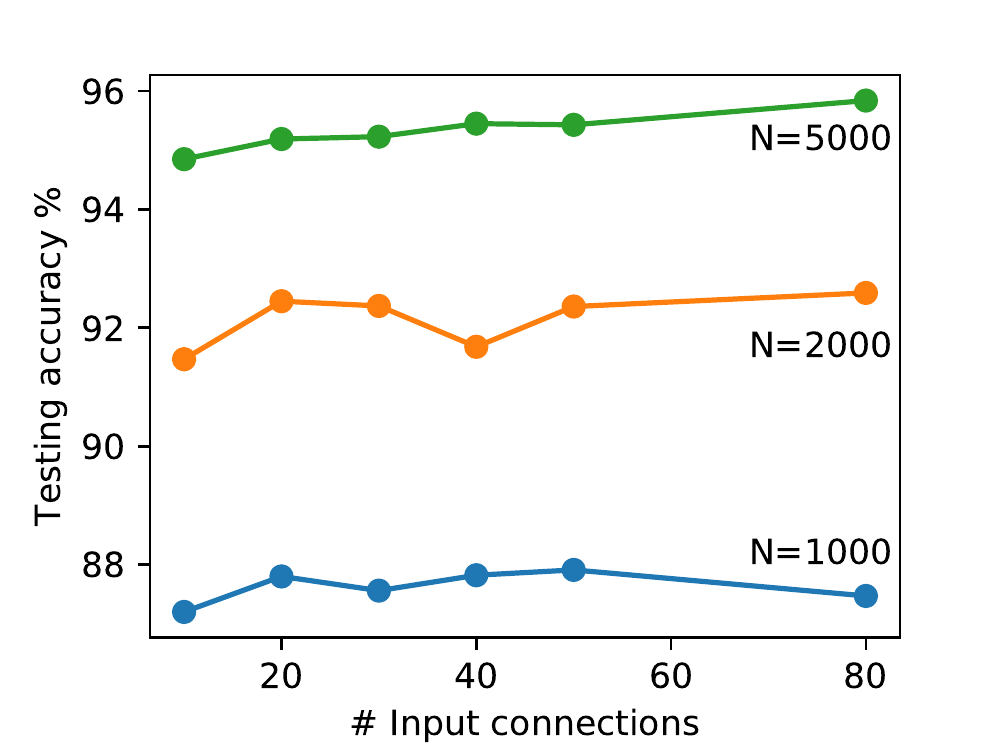}
\caption{MNIST testing accuracy for a mushroom body 
fan-out fan-in type of architecture as a function of both the number of Kenyon
cells $N$ and the number of input connections $n_c$ }
\label{fig_kcmnist}
\end{figure}

In order to understand the role of each of them in pattern recognition tasks, we 
have trained a mushroom body type of network against the MNIST dataset.
 The results presented in Figure \ref{fig_kcmnist} show that the number of Kenyon cells, that is,
the dimensionality of the fan-out parameter,
is the critical variable controlling testing accuracy. In contrast, the number of random connections $n_c$ has a second
order influence. Note that the results obtained in Figure \ref{fig_kcmnist} were obtained using the same metalearning parameters
of the stochastic gradient descent method using ramdom fan-out connectivity.
Consequently, there is the possibility that higher accuracies could be obtained in some
cases through a better choice of these parameters. In the case of 5000 Kenyon cells, the mushroom body type of architecture
was able to achieve 95\% testing accuracies.

\begin{figure}[!tb]
\centering
\includegraphics[width=3in]{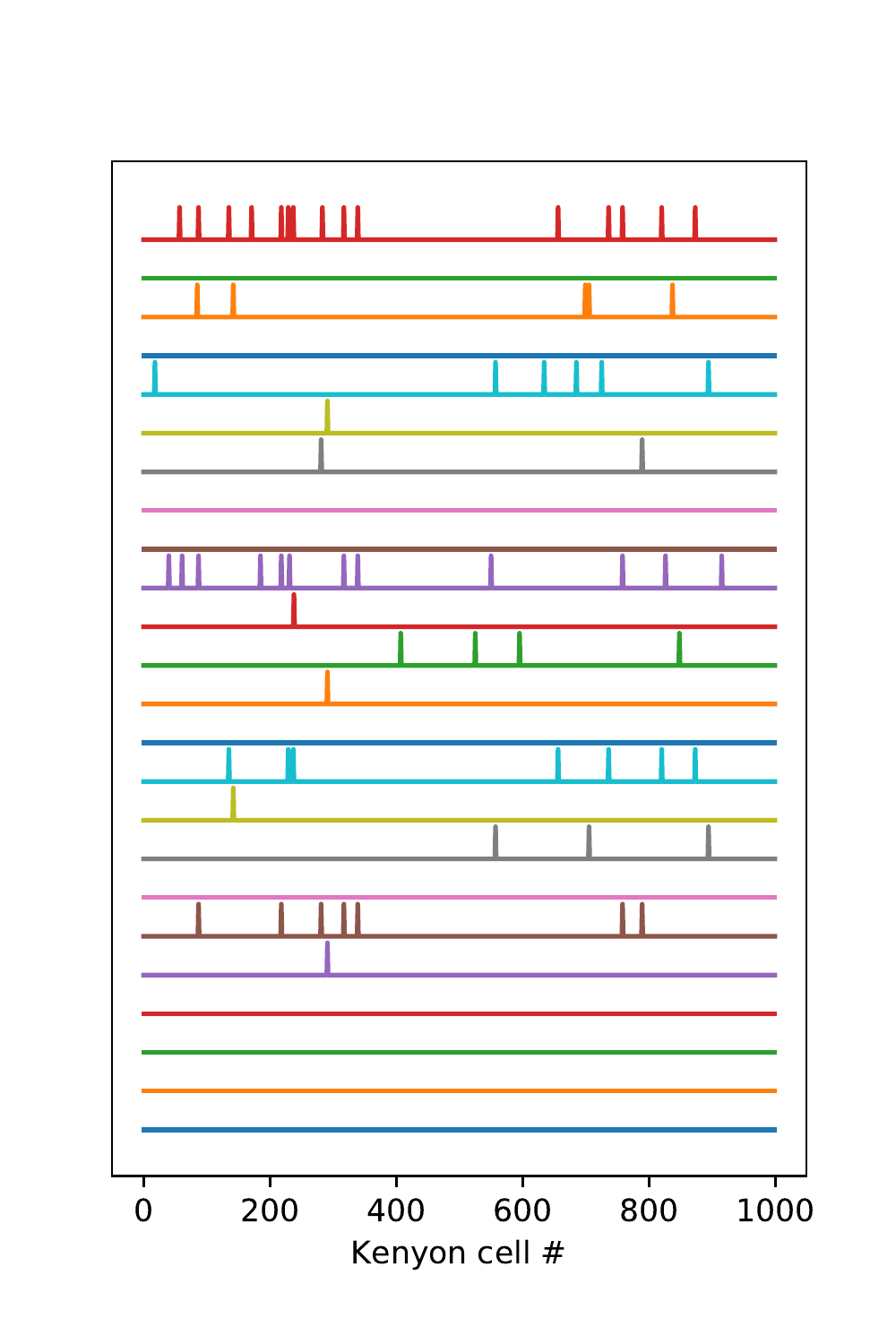}
\caption{Neuron activity as a function of Kenyon cell \# over 24 intervals after receiving a MNIST digit codified using a
time delay + spike train encoding input.
The presence of an inhibitory neuron enhances the sparsity of the response both in time and over the range of active Kenyon cells.
No recurrent interactions among Kenyon cells was considered for this example.}
\label{fig_kc_example}
\end{figure}

Similar to the studies shown in the prior section, we can implement the same type of network using spiking neurons. 
These neurons were modulated by an inhibitory network
as explained in the prior section. When subject to a spike train encoding of input data it results on a complex sparse
activity pattern due to the strong inhibitory component. One such example is shown in Figure \ref{fig_kc_example} for one of the images of the MNIST training set for a network composed of 1000 Kenyon cells, each receiving the input of 70 randomly selected
data streams.

\begin{figure}[!tb]
\centering
\includegraphics[width=3in]{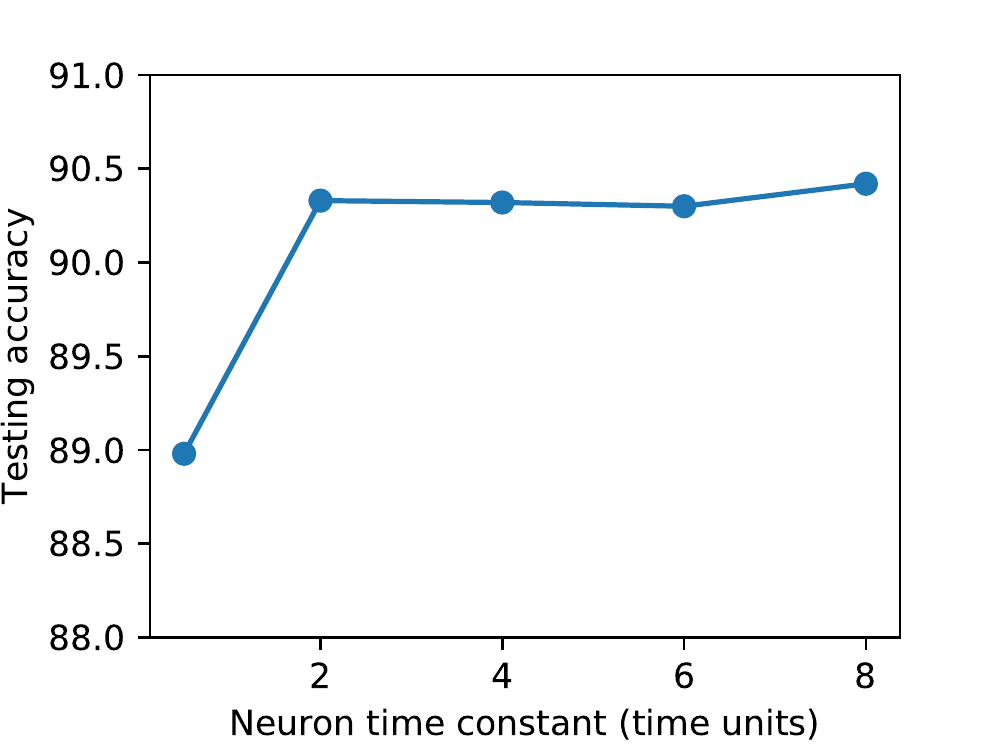}
\caption{MNIST accuracy of a mushroom body-inspired network containing 1000 Kenyon cells, each
receiving 70 randomly selected inputs. Kenyon cell rates over 24 periods are used to fit an output layer of spiking
neurons. Results are shown for output neurons with different neuron time constants $\tau_m$  }
\label{fig_kc_mnist}
\end{figure}

When the output of the Kenyon cells is fed to an output layer of spiking neurons using a rate encoding, we can use stochastic
gradient descent methods to optimize the response of the output layer. The resulting testing accuracy again exceeds
 89\%, as shown in Figure \ref{fig_kc_mnist}. When we combine a time encoding with sparse encoding, the joint accuracy of the combined system is 95\%. The
fast evaluation through time encoding is carried out in only 8 time cycles, whereas for the mushroom body we considered
the output rate after 24 cycles.

 \subsection{The mushroom body architecture provides a substrate for online associative learning and hypothesis-driven perception}
\label{sec_modulation}

A key advantage of the fan-out fan-in architecture
is that it provides a substrate for on-line associative learning. So far all the approaches used in Section \ref{sec_encodings} rely on
the use of stochastic gradient descent methods to train the network for a particular task. In the fan-out fan-in architecture
we can use  modulatory signals to establish associations between certain activity patterns and a desired output.

\begin{figure}[!tb]
\centering
\includegraphics[width=3in]{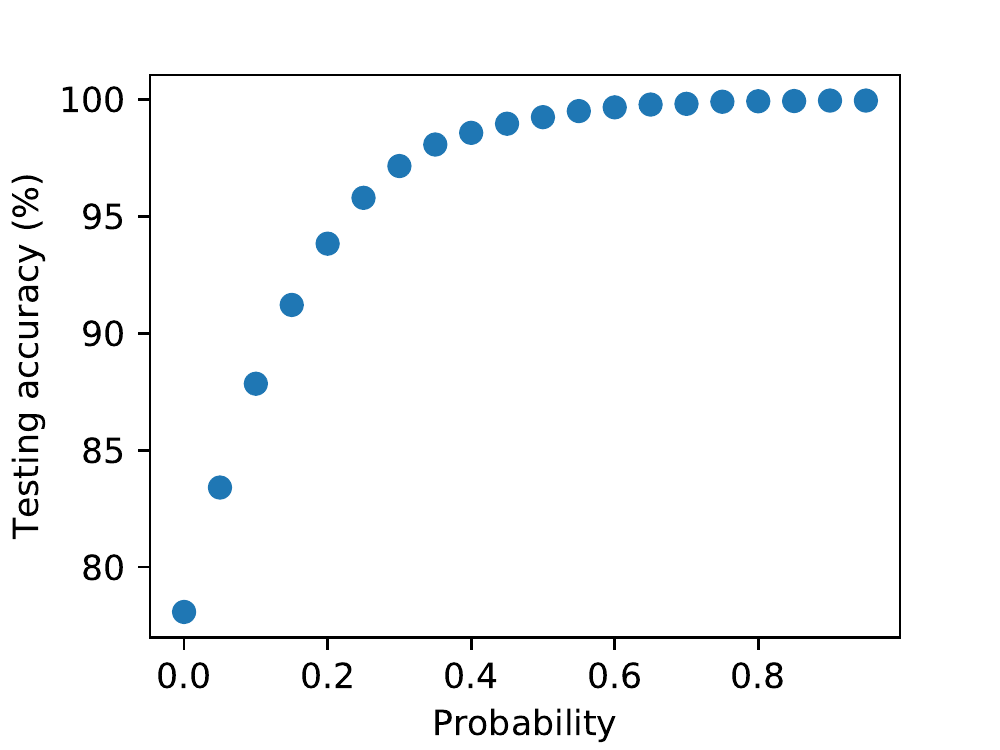}
\caption{Modulation of MNIST testing accuracy on a mushroom body type of architecture as a function of prior knowledge }
\label{fig_prob}
\end{figure}

One such example is the redirection of the output of the fast component to help the Kenyon cell
learn patterns online. This provides a simple way of loading information into the slow, plastic component only as it is
presented or required by the system. Another example is the use of the fast component to modulate a probabilistic interpretation
of the input. One such example is shown in Figure \ref{fig_prob}, where the \emph{a priori} probabilistic assumption of the input
increases the ability of the system to enhance the testing accuracy of a fan out architecture optimized using modulated 
plasticity.

\begin{figure}[!tb]
\centering
\includegraphics[width=3.75in]{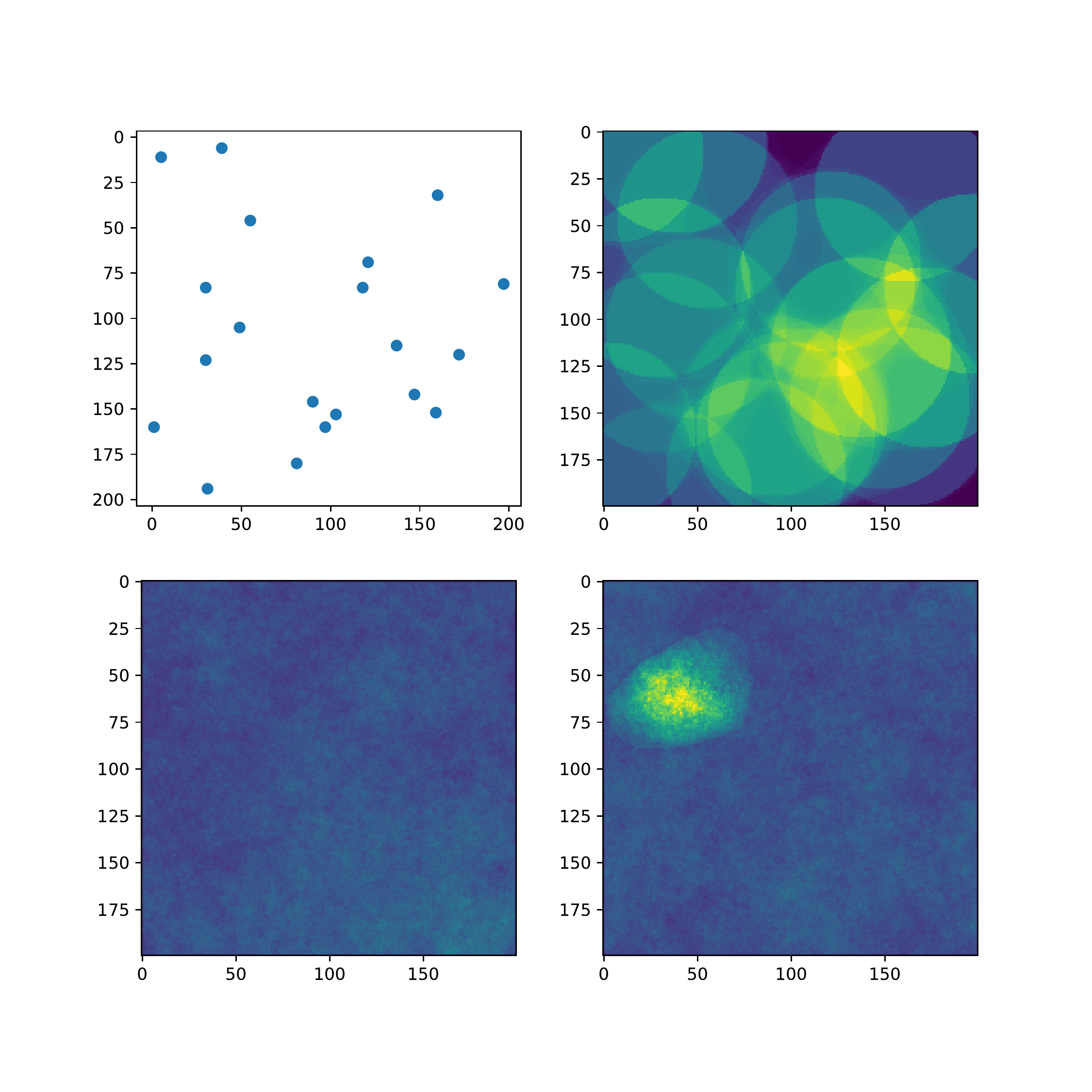}
\caption{Aggregated dwell times of an agent carrying out a random walk over a 2D map.  Top left: spatial distribution of
source points providing the input of the system. Top right: fraction of action Kenyon cells as a function of the position.
Bottom left: Dwell time distribution in absence of reinforcement. Bottom right: Dwell time distribution after on-line reinforcement of
the central position of the map, showing an increase of dwell time close of the excitation area in the top left quadrant.}
\label{fig_dwell}
\end{figure}

A more compelling example is to use this modulatory component to change the behavioral output of a system. 
Here we show an agent carrying out a random walk on a 2D space receiving a series of spatially correlated input data.
For this particular
example we consider a series of point sources randomly distributed over a 2D map (Figure \ref{fig_dwell}).
These sources emit a particular
combination of input data. These can be interpreted as either specific set of pulses or frequencies, in the case of radio or
visual navigation, or it can be series of molecular signals. We consider that the intensity of these pulses decay with distance following
a gaussian profile. 

The network is composed of 20 input neurons, 200 Kenyon cells and one output neuron signaling the valence of the input. The agent
carries out a random walk over the map with a dwell time that is modulated by the valence associated to the input
(Figure \ref{fig_dwell}). A positive valence leads to longer dwell times during the random walk. 
In absence of modulatory input, the dwell time is homogeneously distributed on the map. In order to train the system to spend
more time at the center of the map, we introduced a modulatory signal that strengthen the positive response with a certain
probability $p_0$ whenever the agent is within a certain region.  A feedback path ensures that,
once the system has learned the desired response, the modulatory neuron becomes inactive and the
synaptic weights are no longer updated.

\section{Discussion and conclusions}

The results obtained allow us to extract some useful lessons from a hardware perspective:  first, they
 indicate that the learning capabilities of different types of encodings in spiking neural networks do
not significantly differ from the traditional non-spiking counterparts. Moreover, we showed how both using random
spike rate and deterministic spike train encodings it was possible to achieve similar MNIST testing accuracies in as 
few as eight time steps. This suggests that results obtained in the literature using random spike trains could be reproduced
using deterministic spike rates in hardware as long as the characteristic response time of the neuron is larger than a
single time step. From a sensing and processing point of view, this implies that spiking systems could process information
in a few clock cycles per layer even when a spike rate encoding is used.

Second, the exploration of sparse coding in fan-out fan-in architectures inspired in the insect's mushroom body shows 
that these systems achieve both good testing accuracy and the ability to carry out online learning. An interesting feature
of such systems is that, similar to liquid state machines, learning only takes place in the fan-in region. 
\cite{Maass_liquidstatemachine_2002}
Consequently, this type of architecture is amenable to a modular design, 
where the online/dynamic plasticity component is contained within a separate module. They are also compatible with existing
hardware approaches.

Finally, modulatory learning provides an effective approach for implementing
online associative learning that is conditional to an internal state or external command. The central idea is that the synaptic
strength between two neurons is mediated by a third neuron, bringing together the logic and learning aspects of
neural systems. While the focus of this work was not on devices,
we have emulated such modulatory learning using both memristors and CMOS-based approaches.

\section*{Acknowledgements}

This material is based upon work supported by Laboratory Directed Research and
Development (LDRD) funding from Argonne National Laboratory, provided by the Director,
Office of Science, of the U.S. Department of Energy under Contract No. DE-AC02-06CH11357

Approved for public release: distribution is unlimited. Approval ID: ANL-2017-141245




%


\end{document}